%% file: ReviewTemplate.tex
\documentclass[10pt,twocolumn,letterpaper]{article}

\usepackage{cvpr}              % To produce the CAMERA-READY version
\usepackage{graphicx}
\usepackage{amsmath}
\usepackage{amssymb}
\usepackage{float}
\usepackage{booktabs}
\usepackage{multirow}
\usepackage{url}
\usepackage{placeins}
\usepackage{booktabs}
\usepackage{appendix}
\input{math_commands.tex}

\newcommand{\Ocal}{\mathcal{O}}
\newcommand{\Acal}{\mathcal{A}}

\usepackage[pagebackref,breaklinks,colorlinks]{hyperref}

\usepackage[capitalize]{cleveref}
\crefname{section}{Sec.}{Secs.}
\Crefname{section}{Section}{Sections}
\Crefname{table}{Table}{Tables}
\crefname{table}{Tab.}{Tabs.}

\def\cvprPaperID{6495} % *** Enter the CVPR Paper ID here
\def\confName{CVPR}
\def\confYear{2022}

\begin{document}

\title{MuKEA: Multimodal Knowledge Extraction and Accumulation for Knowledge-based Visual Question Answering}
%\title{Gram: Instantiated Fine-Grained Multimodal Knowledge Extraction for Open-Domain Knowledge-Based VQA}

\author{\textbf{Yang Ding$^{1,2}$, Jing Yu$^{1,2*}$, Bang Liu$^{3,4\dag}$, Yue Hu$^{1,2}$, Mingxin Cui$^{1,2}$, Qi Wu$^{5}$}\\
$^{1}$Institute of Information Engineering, Chinese Academy of Sciences, Beijing, China\\
$^{2}$School of Cyber Security, University of Chinese Academy of Sciences, Beijing, China\\
$^{3}$Universit{\'e} de Montr{\'e}al, Canada\quad\quad
$^{4}$Mila - Quebec AI Institute, Canada\\
$^{5}$University of Adelaide, Australia\\
{\tt\small \{dingyang, yujing02, huyue, cuimingxin\}@iie.ac.cn, bang.liu@umontreal.ca}\\ {\tt\small qi.wu01@adelaide.edu.au}}
\maketitle
\renewcommand{\thefootnote}{\fnsymbol{footnote}}
\footnotetext[1]{Corresponding author.}
\footnotetext[2]{Canada CIFAR AI Chair.}
\renewcommand{\thefootnote}{\arabic{footnote}}
%%%%%%%%% ABSTRACT
\begin{abstract}
Knowledge-based visual question answering requires the ability of associating  external knowledge for open-ended cross-modal scene understanding. One limitation of existing solutions is that they capture relevant knowledge from text-only knowledge bases, which merely contain facts expressed by first-order predicates or language descriptions while lacking complex but indispensable multimodal knowledge for visual understanding. How to construct vision-relevant and explainable multimodal knowledge for the VQA scenario has been less studied. In this paper, we propose MuKEA to represent multimodal knowledge by an explicit triplet to correlate visual objects and fact answers with implicit relations. To bridge the heterogeneous gap, we propose three objective losses to learn the triplet representations from  complementary views: embedding structure, topological relation and semantic space. By adopting a pre-training and fine-tuning learning strategy, both basic and domain-specific multimodal knowledge are progressively accumulated for answer prediction. We outperform the state-of-the-art by 3.35$\%$ and 6.08$\%$ respectively on two challenging knowledge-required datasets: OK-VQA and KRVQA. Experimental results prove the complementary benefits of the multimodal knowledge with existing knowledge bases and the advantages of our end-to-end framework over the existing pipeline methods. The code is available at \url{https://github.com/AndersonStra/MuKEA}.

\end{abstract}

%%%%%%%%% BODY TEXT
\section{Introduction}
\label{sec:intro}
\begin{figure}[t]
\setlength{\abovecaptionskip}{-5pt}
\setlength{\belowcaptionskip}{-15pt}
\begin{center}
\includegraphics[width=1.0\linewidth]{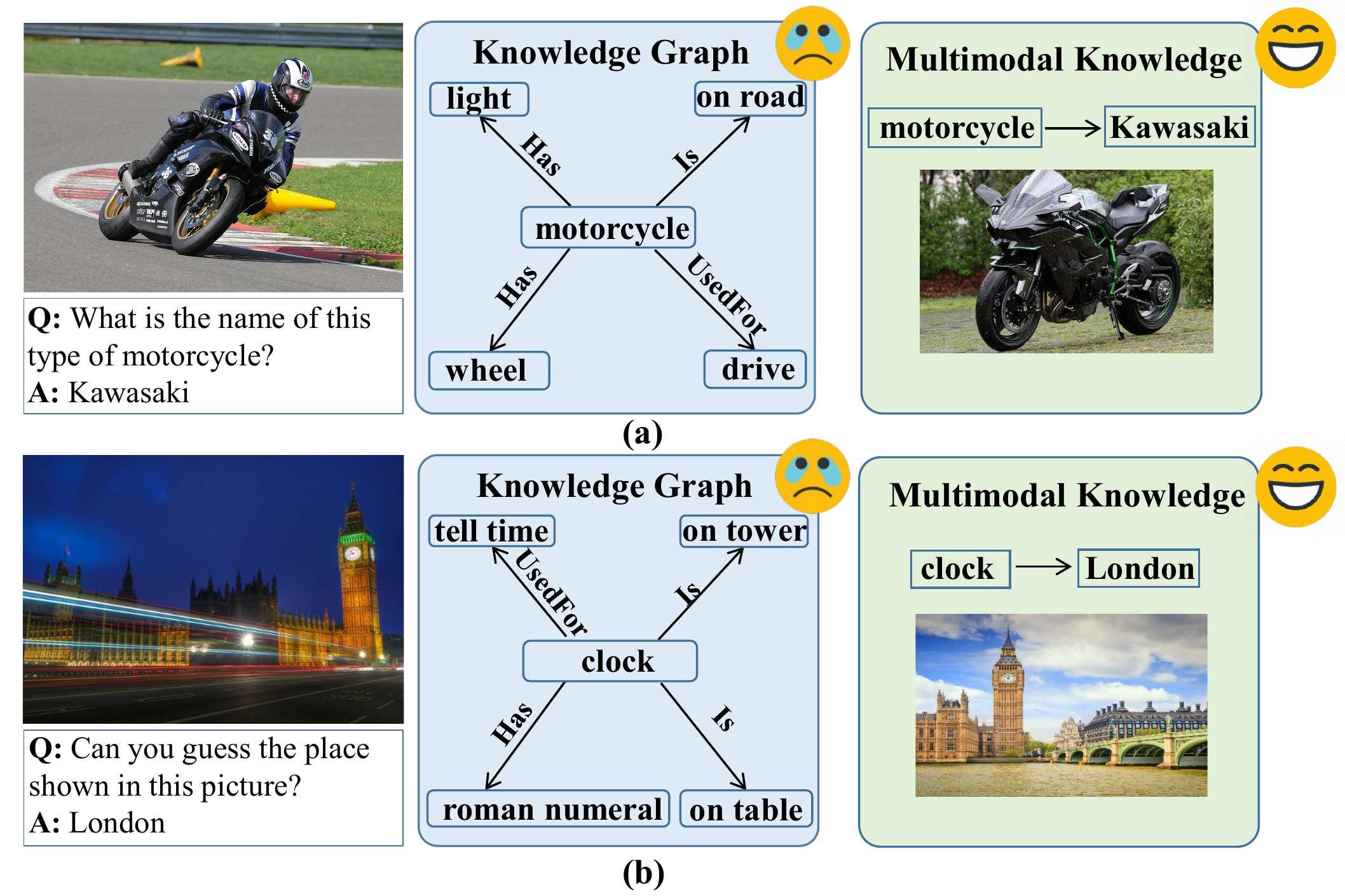}
\end{center}
\caption{An illustration of our motivation. Compared with rigid facts in the knowledge graph, multimodal knowledge for depicting complex and inexpressible facts is indispensable in both open-ended object understanding (a) and scene understanding (b). }
\label{fig:intro}
\end{figure}

Visual Question Answering based on external Knowledge Bases (KB-VQA) \cite{wang2017explicit} requires an AI agent to answer a question by incorporating knowledge about the world beyond what the question and the image contains. Despite the great success in VQA tasks \cite{han2021focal,han2020interpretable}, KB-VQA is more challenging for models to achieve human-like ability of open-ended cross-modal scene understanding  associating with external knowledge.  Therefore, how to appropriately represent and leverage knowledge in such cross-modal scenario becomes a core problem of KB-VQA.

Most of recent works \cite{garderes2020conceptbert, marino2021krisp, DBLP:conf/ijcai/ZhuYWS0W20} focus on capturing relevant knowledge from structured knowledge graphs, such as ConceptNet \cite{liu2004conceptnet} and DBpedia \cite{auer2007dbpedia}, or unstructured/semi-structured knowledge, like Wikipedia \cite{wiki} and Visual Genome \cite{krishna2017visual}. Though these knowledge bases provide high-quality  knowledge by large-scale human annotations, the information is generally limited to the definite facts that can be explicitly expressed by natural language or simple triplets with first-order predicate. Therefore, such knowledge bases are quite difficult to represent high-order predicate and multimodal knowledge, which is essential for human to tackle complex problems. Considering the question in Figure \ref{fig:intro}(a), the agent needs visual knowledge of motorcycle appearance in each
brand to identify the given motorcycle, but the knowledge graph lacks of such instantiated information. Besides object understanding, implicit visual knowledge in mind mostly dominate over the rigid facts when humans are asked for simple scene discrimination like the question `Can you guess the place?' in Figure \ref{fig:intro}(b). \textit{How to represent and accumulate the complex multimodal knowledge in the VQA scenario while maintaining the advantages of traditional knowledge graph in explainable reasoning is an essential but less studied problem.}

Current progress \cite{sun2020multi, pezeshkpour2018embedding, li2020gaia} in emerging multimodal knowledge graph aims to correlate visual content with textual facts to form the augmented knowledge graph. The typical solutions can be divided into two categories: parsing images and texts to structured representations and grounding event/entities across modalities \cite{li2020gaia, xie2021construction, kannan2020multimodal}, or simply aligning the entities in existing knowledge graphs with  related images \cite{pezeshkpour2018embedding, sun2020multi}. However, such multimodal knowledge graphs in essence still represent knowledge via the first-order predicate, which fails to model the high-order complex relationships such as the relationship between `clock' and `London' in Figure \ref{fig:intro}(b).  

In this paper, we propose a novel \textit{\textbf{Mu}ltimodal \textbf{K}nowledge \textbf{E}xtraction and \textbf{A}ccumulation} framework (MuKEA) for KB-VQA task. Independent of existing knowledge bases, the core mechanism behind MuKEA is to accumulate  multimodal knowledge with complex relationships from observation of VQA samples, and perform explainable reasoning based on the self-accumulated knowledge. To this end, we first propose a novel schema to represent multimodal knowledge unit by an explicit triplet, where the visual objects referred by the question are embedded in the head entity,
% where the visual objects and corresponding context referred by the question are embedded in the head entity, the fact knowledge serving as the answer of the question about the visual content is kept in the tail entity, and the complex visual and semantic causality between the head and the tail is expressed by the relation. 
the embedding of the fact answer is kept in the tail entity, and the implicit relation between the head and the tail is expressed by the relation. 
We propose three objective loss functions to learn the representations of the triplets from coarse to fine by contrasting positive and negative triplets, aligning ground-truth triplets, and refining entity representations. A pre-training and fine-tuning learning strategy is then proposed to progressively accumulate multimodal knowledge from both out-domain and in-domain VQA samples for explainable reasoning. 

The main contributions of this work are as follows: 

(1) We propose an end-to-end multimodal knowledge %extraction and 
representation learning framework, which first models the  inexpressible multimodal facts by explicit triplets and provides complementary knowledge with the existing knowledge graphs and unstructured knowledge bases. 

(2) We exploit a pre-training and fine-tuning strategy to accumulate both out-domain %basic knowledge 
and in-domain %specific 
knowledge to form a neural multimodal knowledge base. It supports automatic knowledge association and answer prediction, which gets rid of the cascading error in existing `knowledge retrieve and read' pipeline \cite{DBLP:conf/ijcai/ZhuYWS0W20, marino2021krisp}.
% and the choice limitation in typical answer classification solutions \cite{garderes2020conceptbert, marino2019ok}. 

% (3) Our model with strong generalization ability sets a new state-of-the-art on two challenging KB-VQA datasets: OK-VQA \cite{marino2019ok} and KRVQA \cite{cao2021knowledge}, significantly outperforming prior works including the pre-trained models. The good performance can be well explained by visualizing the relevant multimodal knowledge triplets explicitly.
(3) Our model with strong generalization ability outperforms the state-of-the-art models by 3.35$\%$ and 6.08$\%$ respectively on two challenging KB-VQA datasets: OK-VQA \cite{marino2019ok} and KRVQA \cite{cao2021knowledge}. %, including the pre-trained models. 
The good performance can be well explained by visualizing the relevant multimodal knowledge triplets explicitly.

\section{Related Work}

\noindent\textbf{Knowledge-based Visual Question Answering.}\quad Most of recent works are based on `knowledge retrieve and read' pipeline which requires highly-relevant knowledge to support knowledge reasoning. Structured knowledge based methods like \cite{garderes2020conceptbert} is based on ConceptNet \cite{liu2004conceptnet} to introduce knowledge in the form of triplet with first-order predicate. Unstructured knowledge based methods \cite{marino2019ok} retrieve knowledge from Wikipedia \cite{wiki} and encode relevant text in a memory network for further reasoning. However, knowledge described in nature language lacks visual information to 
assist cross-modal understanding. For the above challenge, \cite{zheng2021km4} augments the knowledge graph YAGO  \cite{rebele2016yago} with related images to serve as multimodal knowledge. However, such graph in essence still represents knowledge via the first-order predicate. To go one step further, we extract multimodal information to represent high-order complex relations and represent multimodal knowledge by  explicit triplets for explainable reasoning.

From the view of model framework, most of recent works are based on the `retrieve and read' pipeline, which first retrieve the relevant facts from knowledge bases and then perform explicit reasoning on the knowledge graph \cite{narasimhan2018out,wang2017explicit,wang2017fvqa},  or fusing the implicit knowledge embedding with image and questions for answer classification \cite{garderes2020conceptbert,marino2019ok,li2020boosting,wu2016ask}. All of these methods rely on object labels to retrieve external knowledge, which inevitably introduces irrelevant knowledge and leads to cascading error. There are also end-to-end methods based on implicit knowledge like pre-trained models \cite{tan2019lxmert, lu2019vilbert,marino2021krisp}. However, such implicit knowledge mainly captures the co-occurrence of image-question-answer triplet instead of explainable and refined knowledge. In this paper, we propose an  end-to-end  multimodal  knowledge extraction and accumulation framework with interpretable triplet knowledge.

\noindent\textbf{Multimodal Knowledge Graph.}\quad The emerging multimodal knowledge graph works \cite{sun2020multi, pezeshkpour2018embedding, li2020gaia} aim to correlate visual content with textual facts to form the augmented knowledge graph. One typical solution parses images and texts to structured representations first and grounds event/entities across modalities. The key problem lies in intra-modal relation extraction and cross-modal entity linking. Specifically, \cite{li2020gaia, nian2017multi} learn knowledge from structured textual and visual data and maintain the triplet structure for entity alignment. \cite{kannan2020multimodal} utilizes RDF \cite{manola2004rdf} knowledge graphs to represent multimodal information based on 
graph alignment and lacks multimodal correlation as well. Another kind of solutions directly links the entities in existing knowledge graphs with relevant images.  \cite{pezeshkpour2018embedding} adds images to expand the entity representation in YAGO \cite{rebele2016yago}. However, all of these methods in essence still represent knowledge via the first-order predicate described by nature language, which fails to model the high-order complex relationships.

%------------------------------------------------------------------------

\begin{figure*}[htb]
\setlength{\abovecaptionskip}{-5pt}
\setlength{\belowcaptionskip}{-12pt}
\begin{center}
\includegraphics[width=1.0\linewidth]{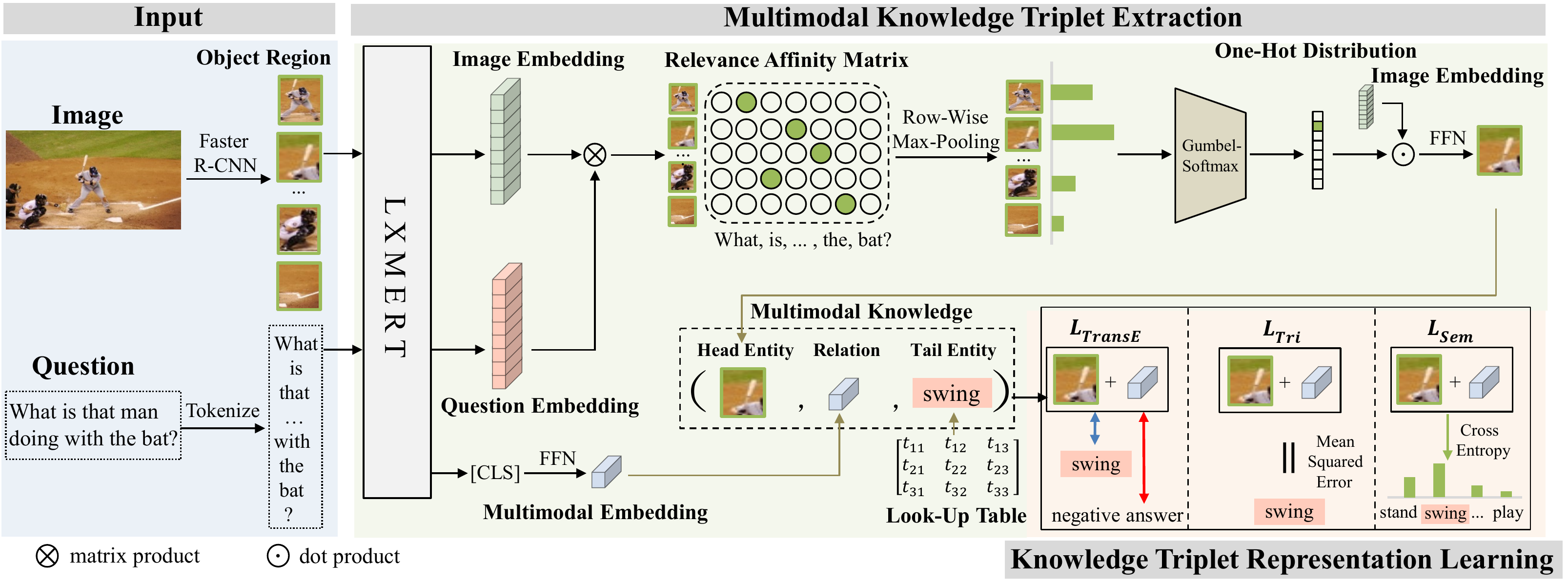}
\end{center}
\caption{An overview of our model. The model contains two modules: Multimodal Knowledge Triplet Extraction aims to  extract  multimodal  knowledge triplets from samples and Knowledge Triplet Representation Learning aims to unifiedly learn the triplet representation.}
\label{fig:model}
\end{figure*}

\section{Methodology}
% knowledge-based VQA intends to ground answer based on external linguistic knowledge bases, In order to improve the recall of relevant knowledge, previous methods usually retrieve knowledge from multiple knowledge bases, however, Knowledge stored in text form is not enough to support answering visual questions, and  knowledge retrieved directly for the question and image is often noisy and not useful for predicting the correct answer. 

Given an image $I$ and a question $Q$, the KB-VQA task aims to predict an answer $A$ supported by additional knowledge beyond the given visual and textual content. We accumulate triplet-formed multimodal knowledge  to serve as the external knowledge and directly infer the answer in an end-to-end mode. Figure \ref{fig:model} gives detailed illustration of our model. We first introduce a novel schema of extracting multimodal knowledge triplets from unstructured image-question-answer samples based on the pre-trained vision-language model. Then we propose three objective losses to learn the triplet embeddings that accurately depict question-attended visual content (head embeddings), question-desired fact answer (tail embeddings), and the implicit relation between the two (relation embeddings). By training with both out-domain and in-domain data, our model accumulates a wide range of multimodal knowledge and  associates the optimal fact for answer prediction.

\subsection{Multimodal Knowledge Triplet Extraction}
\label{sec:mke}

In the VQA scenario, we define the complex and inexpressible facts as multimodal knowledge in the form of triplet, \textit{i.e.} $(h, r, t)$, where $h$ contains visual content in the image focused by the question, $t$ is a representation of the answer given the question-image pair, and $r$ depicts the implicit relationship between $h$ and $t$ containing multimodal information. The triplet construction process mainly consists of the following four parts:

\noindent\textbf{Image  and Question Encoding.}\quad Since the pre-trained vision-language models are strong at modeling the intra-modal and cross-modal implicit correlations, we first utilize the pre-trained model LXMERT \cite{tan2019lxmert} to encode the question and image for further multimodal knowledge triplet extraction. We apply Faster R-CNN \cite{ren2015faster}  to detect a set of objects
$\Ocal = \{o_i\}_{i=1}^K$ ($K$ = 36) in $I$ and represent 
each object $o_i$ by a visual feature vector $\bm{f}_i \in \R^{d_f}$ ($d_f$ = 2048) and a spatial feature vector $\bm{b}_i \in \R^{d_b}(d_b = 4)$ as in \cite{DBLP:conf/ijcai/ZhuYWS0W20}. We tokenize a question $Q$ using WordPiece \cite{wu2016google} and obtain a sequence of $D$ tokens.
We feed the visual features $\{\bm{f}_i\}_{i=1}^K$ and $\{\bm{b}_i\}_{i=1}^K$, and question tokens into the pre-trained LXMERT, obtaining the visual embeddings of $\Ocal$ denoted as $\bm{V} \in \R^{K \times d_v}$ ($d_v$ = 768) and the token embeddings denoted as $\bm{Q} \in \R^{D \times d_v}$.

\noindent\textbf{Head Entity Extraction.}\quad
We define the head entity as the visual object and its context in the image that is most relevant to the question. %thus we leverage hard attention mechanism to adaptively select the strong semantically relevant image regions, the head entity $h$ can be obtained as below:
%\begin{equation}\label{1}
%h=\mathop{\arg\max}_{h \in O} P_{\Theta}(h|I, Q)
%\end{equation}
%where $h$ is a object in the image, $\Theta$ is the parameters of the model P that needs to be trained
To this end, we firstly evaluate the relevance of each object in the image to each token in the question by computing the question-guided object-question relevance affinity matrix $\bm{A}$ as: 
\begin{equation}\label{simi}
\mA= (\textbf{W}_1 \bm{Q})^T(\textbf{W}_2 \bm{V})
\end{equation}
where $\textbf{W}_1$ and $\textbf{W}_2$ are learned parameters.

Under the guidance of the relevance affinity matrix, we then select one object in $\Ocal$ as the most relevant visual content to the question. Since the LXMERT models the implicit correlations among all the objects, it is noteworthy that the selected question-centric object already contains its context information, which provides indispensable clues for answering  questions that involve multiple objects. Specifically we compute the row-wise max-pooling on $\bm{A}$ to evaluate the relevance of each object $o_i$ to the question as:
\begin{equation}\label{max}
\va^{v-q}_i= \mathop{\max}_{j}\mA_{i,j}
\end{equation}
Then hard attention instead soft attention is applied to select the most relevant object as the head entity based on $\{\va^{v-q}_i\}_{i=1}^K$. Compared with soft attention, hard attention provides more stable and explainable visual content for multimodal knowledge representation, which is also easier for combining with exiting knowledge graph by entity linking. Here we conduct Gumbel-Softmax \cite{jang2016categorical} to obtain the approximate one-hot categorical distribution. The attention weight for object $o_i$ is computed as:
\begin{equation}\label{2}
\alpha_i = \frac{\exp((\log(\va^{v-q}_i)+g_i)/\tau)}{\sum_{j=1}^{K}\exp((\log(\va^{v-q}_j)+g_j)/\tau)}
\end{equation}
where $\{g_i\}_{i=1}^K$ are \textit{i.i.d.} samples drawn from Gumbel(0,1)\footnote{The Gumbel (0,1) distribution can be sampled using inverse transform sampling by drawing $u\sim$ Uniform(0,1) and computing $g=-\log(-\log(u)))$}, and $\tau$ is the softmax temperature. Finally, we gather the question-centric object information and obtain the head entity representation $\bm{h}$ as:
\begin{equation}\label{3}
\vh = {\rm FFN}(\sum_{i=1}^{K}\alpha_i\bm{v}_{i}) 
\end{equation}
where $\bm{v}_{i}\in \bm{V}$ and FFN denotes a feed-forward network that contains two fully connected layers.

\iffalse
and max pooling the matrix to obtain the maximum similarity score between each object and the question, then we utilize Gumbel-Softmax \cite{jang2016categorical} to sample the one-hot categorical distribution vector $\alpha \in \R^{K \times 1}$ and handle the discreteness problem of hard attention mechanism, finally the representation of head entity $\vh$ can be calculated as below:
\begin{equation}\label{simi}
\mA= (\mW_1 \textbf{q})^T(\mW_2 \textbf{v})
\end{equation}
\begin{equation}\label{max}
\va^v_i= \mathop{\max}_{j}\mA_{i,j}
\end{equation}
\begin{equation}\label{2}
\alpha_i = \frac{\exp((\log(\va^v_i)+g_i)/\tau)}{\sum_{j=1}^{k}\exp((\log(\va^v_j)+g_j)/\tau)}
\end{equation}
\begin{equation}\label{3}
\vh = {\rm FFN}(\alpha^T \textbf{v}) 
\end{equation}
where $\mW_1$ and $\mW_2$ are learned parameters, $\mA$ is the affinity matrix of the two modalities, $g_1...g_k$ are i.i.d samples drawn from Gumbel(0,1)\footnote{The Gumbel (0,1) distribution can be sampled using inverse transform sampling by drawing $u\sim$ Uniform(0,1) and computing $g=-\log(-\log(u)))$}, $\tau$ is the softmax temperature, FFN denotes a feed-forward network that contains two FC layers (specifically, FC-ReLU-FC).

We use Gumbel-Softmax as a continuous, differentiable approximation to $\arg\max$, so the semantic label of the head entity is retained. 
\fi

\noindent\textbf{Relation Extraction.}\quad Different from the relation in traditional knowledge graph that depicts the first-order predicate independent of specific visual scenario, we define the relation in multimodal knowledge as the complex implicit relation between the observed instantiated object and the corresponding fact answer. Since LXMERT captures the implicit correlations between the image and the question via the self-attention mechanism in the hierarchical transformers, we extract the cross-modal representation from the [CLS] token, and feed it into a FFN layer to obtain the relation embedding, which is denoted as $\vr$.

\noindent\textbf{Tail Entity Extraction.}\quad
We define the tail entity as the answer in an image-question-answer sample, which reveals a specific aspect of facts regarding to the visual object referred by the question.  In the training stage, we set ground truth answer as the tail entity to learn its representation $\bm{t}$ from scratch (details in Section \ref{sec:krl}). In the inference stage, we define the KB-VQA task as a multimodal knowledge graph completion problem and globally assess the knowledge in our neural multimodal knowledge base to predict the optimal tail entity as the answer (details in Section \ref{ssec:prediction}).

\subsection{Knowledge Triplet Representation Learning}
\label{sec:krl}
Since each component within a triplet contains modality-different and semantic-specific information, we propose three loss functions to unifiedly learn the triplet representation in order to bridge the heterogeneous gap as well as semantic gap. The three losses constrain the triplet representation from complementary views: the \textit{Triplet TransE Loss} preserves the embedding structure by contrasting positive and negative triplets. The \textit{Triplet Consistency Loss} further forces the three embeddings within a triplet to keep the strict topological relation, and the \textit{Semantic Consistency Loss} maps the embeddings into a common semantic space for fair comparison among multimodal content.

%After knowledge extraction as described in Section \ref{sec:mke}, in this section, we introduce the knowledge representation learning method for multimodal knowledge.

\noindent\textbf{Triplet TransE Loss.}\quad
Inspired by the knowledge embedding method TransE \cite{bordes2013translating} in the traditional knowledge graph field, we apply TransE-like objective loss as a structure-preserving constraint in our multimodal scenario. Given an image-question pair, let $\Acal^+$ and $\Acal^-$ denote its sets of correct (positive) and incorrect (negative) answers, respectively. Let $\bm{h}$ and $\bm{r}$ denote the corresponding extracted head and tail entity representations. We want the distance between $\bm{h}+\bm{r}$ and each positive tail $\bm{t}^+\in \Acal^+$ to be smaller than the distance between $\bm{h}+\bm{r}$ and each negative tail $\bm{t}^-\in \Acal^-$ by a certain margin $\gamma$:
\begin{equation}\label{4}
\mathcal{L}_{\rm TransE} = \sum_{\bm{t}^+\in \Acal^+}\sum_{\bm{t}^-\in \Acal^-}[\gamma + {\rm d}(\bm{h}+\bm{r},\bm{t}^+) - {\rm d}(\bm{h}+\bm{r},\bm{t}^-)]_+
\end{equation}
where $[\cdot]_+\triangleq{\rm max}(0,\cdot)$ and ${\rm d}(\cdot,\cdot)$ denotes the cosine distance following the settings in \cite{luo2019strong}.

\iffalse
For relation embedding in multimodal knowledge has different distribution in each image-question pair instance, training different projection matrices\cite{wang2014knowledge,lin2015learning,ji2015knowledge} for each relation can not be afford, we apply TransE \cite{bordes2013translating} as the knowledge graph embedding method, TransE regards the relation $\vr$ as translation from $\vh$ to $\vt$ for a golden triplet $(h,r,t)$. \cite{luo2019strong} confirmed that the measurement effect of cosine distance is significantly better than Euclidean distance for representation learning, hence we replace the original Euclidean distance in the TransE with cosine distance, then we define the following triplet ranking loss for training:
\begin{equation}\label{4}
\mathcal{L}_{\rm TransE} = \sum_{(h,r,t)\in S}\sum_{(h,r,t')\in S'}[\gamma + d(h+r,t) - d(h+r,t')]_+
\end{equation}
where $\gamma$ is the margin separating golden triplets and negative triplets. $S$ denotes golden triplets set, $S'$ denotes negative triplets set with corrupted tails, and
$[x]_+\triangleq{\rm max}(0,x)$
\fi

\noindent\textbf{Triplet Consistency Loss.}\quad The issue of the above TransE loss is that once the distance between the positive pairs is smaller than the negative pairs by margin $\gamma$ during training, the model will stop learning from the triplet. To further push the embeddings to satisfy the strict topological relation, we apply Mean Squared Error (MSE) criterion to constrain the representations on top of each positive triplet as:
\begin{equation}\label{5}
\mathcal{L}_{\rm Tri} = {\rm MSE}(\bm{h}+\bm{r}, \bm{t}^+)
\end{equation}

\begin{table*}[t]
\centering
\setlength{\tabcolsep}{6mm}{
\begin{tabular}{l|l|l}
\hline
\multicolumn{1}{c|}{\textbf{Method}} & \textbf{Knowledge Resources}          & \textbf{Accuracy} \\ \hline
ArticleNet (AN) \cite{marino2019ok}                      & Wikipedia                             & 5.28                 \\
Q-only \cite{marino2019ok}                               & —                                     & 14.93                \\
BAN \cite{kim2018bilinear}                                  & —                                     & 25.17                \\
\quad+AN \cite{marino2019ok}                             & Wikipedia                             & 25.61                \\
\quad+ KG-AUG \cite{li2020boosting}                       & Wikipedia + ConceptNet                & 26.71                \\
MUTAN \cite{ben2017mutan}                               & —                                     & 26.41                \\
\quad+ AN \cite{marino2019ok}                            & Wikipedia                             & 27.84                \\
Mucko \cite{DBLP:conf/ijcai/ZhuYWS0W20}                                & ConceptNet                            & 29.20                \\
GRUC \cite{DBLP:journals/pr/YuZWZHT20}                                & ConceptNet                            & 29.87                \\
KM$^4$ \cite{zheng2021km4}                               & multimodal knowledge from OK-VQA      & 31.32               \\ \hline
ViLBERT \cite{lu2019vilbert}                               & —                                     & 31.35                \\
LXMERT \cite{tan2019lxmert}                                & —                                   &32.04                   \\
KRISP(w/o mm pre.) \cite{marino2021krisp}                  & DBpedia + ConceptNet + VisualGenome + haspartKB    & 32.31     \\
KRISP(w/ mm pre.) \cite{marino2021krisp}       & DBpedia + ConceptNet + VisualGenome + haspartKB                  & 38.90     \\
ConceptBert \cite{garderes2020conceptbert}                          & ConceptNet                            & 33.66                \\ 
Knowledge is Power \cite{zheng2021knowledge}                  & YAGO3      & 39.24               \\ \hline
% Ours(w/o VQA 2.0 knowledge)                    & multimodal knowledge from OK-VQA                  & 36.35  \\
MuKEA                   & multimodal knowledge from VQA 2.0 and OK-VQA  & \textbf{42.59}               \\ \hline
\end{tabular}}
\caption{State-of-the-art comparison on OK-VQA dataset. The middle column lists the
external knowledge sources, if any, used in each VQA system. The rows in the middle part list the
method based on pre-trained model.}
\label{sota}
\end{table*}

\noindent\textbf{Semantic Consistency Loss.}\quad
We randomly initialize a look-up table of tail entities and learn their representations together with the head and the relation. Each tail entity in the look-up table $\bm{T}$ corresponds to an unique answer in the training VQA samples. To introduce the semantics of answer in tail representation while narrowing the heterogeneous gap between text-formed tail entity and multimodal-formed head entity and relation,
we classify the triplet over the tail vocabulary and force the model to select the ground-truth tail (answer) by the negative log likelihood loss: 
\begin{equation}\label{6}
P(\bm{t}^+) = {\rm softmax}((\bm{T})^T(\bm{h}+\bm{r}))
\end{equation}
\begin{equation}\label{7}
\mathcal{L}_{\rm Sem} = -{\rm log}(P(\bm{t}^+))
\end{equation}
where $P(\bm{t}^+)$ is the predicted probability of ground-truth tail $\bm{t}^+$. In conclusion, our final loss is defined as:
\begin{equation}\label{9}
\mathcal{L}=\mathcal{L}_{\rm TransE}+\mathcal{L}_{\rm Tri}+\mathcal{L}_{\rm Sem}
\end{equation}
% where we found that $\lambda_1=\lambda_2=1$ works reasonably well for all scenarios.
% At inference stage, we define the task as tail entity prediction, we feed the image and question into the network and get the tail entity that best matches the head entity and relation as the predicted answer.
% \begin{equation}\label{8}
% a=\mathop{\arg\min}_{t \in \mathcal{A}} d(h+r,t)
% \end{equation}

\subsection{Knowledge Accumulation and Prediction}
\label{ssec:prediction}
% We pre-train our model on the VQA 2.0 dataset to accumulate prior knowledge via the knowledge extraction and representation learning tasks (Sec. \ref{sec:mke}, \ref{sec:krl}). VQA 2.0 contains about 1.1 million questions, all questions are divided in three categories: Yes/No, Number, and Other. To filter out questions can be answered without common sense knowledge, we only keep Other type questions for pre-training. After filtering, we get 209,126 questions. 
We adopt a two-stage training strategy to accumulate multimodal knowledge progressively: (1) pre-training on the VQA 2.0 dataset \cite{goyal2017making} to accumulate basic visual-dominant knowledge and then (2) fine-tuning on the training data of downstream KB-VQA task to accumulate more complex domain-specific multimodal knowledge. 
% A great amount of questions in VQA 2.0 requires fine-grained visual understanding of object attributes, $e.g.$ categories, colors and materials, which provide basic visual commonsense for understanding more complicated knowledge on top of them. 
All questions in VQA 2.0 are divided into three categories: \textit{Yes/No}, \textit{Number}, and \textit{Other}. 
% Since the first two categories focusing more on the specific visual appearance with less commonsense, 
Since answers in the first two categories can not serve as the fact knowledge,
we only keep \textit{Other} type questions for pre-training. 

In the inference stage, we regard the answer prediction as a multimodal knowledge graph completion problem. Given an image and a question, we feed them into the network and obtain the embeddings of the head entity $\bm{h}_{inf}$ and the relation $\bm{r}_{inf}$. We compute the distance between $\bm{h}_{inf}+\bm{r}_{inf}$ and each tail entity $\bm{t}_i$ in the look-up table $\bm{T}$ and select the tail entity with the minimum distance as:
\begin{equation}\label{8}
\bm{t}_{inf}=\mathop{\arg\min}_{\bm{t}_i \in \bm{T}} {\rm d}(\bm{h}_{inf}+\bm{r}_{inf},\bm{t}_i)
\end{equation}
The answer corresponding to the optimal tail entity $\bm{t}_{inf}$ is selected as the predicted answer.

\section{Experiments}

\begin{table*}[htb]
\centering
\setlength{\tabcolsep}{0.8mm}{
\begin{tabular}{l|ccc|cccc|c|cccc|c}
\hline
\multirow{3}{*}{Method}    & \multicolumn{7}{c|}{KB-not-related}                                                                                  & \multicolumn{5}{c|}{KB-related}                                                           & \multirow{3}{*}{Overall} \\ \cline{2-13}
                           & \multicolumn{3}{c|}{one-step}                    & \multicolumn{4}{c|}{two-step}                                     & \multicolumn{1}{c|}{one-step} & \multicolumn{4}{c|}{two-step}                                   &                          \\ \cline{2-13}
                           & 0              & 1              & 2              & 3              & 4              & 5              & 6              & 2                             & 3           & 4              & 5             & 6              &                          \\ \hline
Q-type \cite{cao2021knowledge}                    & 36.19          & 2.78           & 8.21           & 33.18          & 35.97          & 3.66           & 8.06           & 0.09                          & 0.00          & 0.18           & 0.06          & 0.33           & 8.12                     \\
LSTM \cite{cao2021knowledge}                       & 45.98          & 2.79           & 2.75           & 43.26          & 40.67          & 2.62           & 1.72           & 0.43                          & 0.00          & 0.52           & 1.65          & 0.74           & 8.81                     \\
FiLM \cite{perez2018film}                       & 52.42          & 21.35          & 18.50          & 45.23          & 42.36          & 21.32          & 15.44          & 6.27                          & 5.48          & 4.37           & 4.41          & 7.19           & 16.89                    \\
MFH \cite{yu2018beyond}                       & 43.74          & 28.28          & 27.49          & 38.71          & 36.48          & 20.77          & 21.01          & 12.97                         & 5.10          & 6.05           & 5.02          & 14.38          & 19.55                    \\
UpDn \cite{anderson2018bottom}                       & 56.42          & 29.89          & 28.63          & 49.69          & 43.87          & 24.71          & 21.28          & 11.07                         & 8.16          & 7.09           & 5.37          & 13.97          & 21.85                    \\
MCAN \cite{yu2019deep}                       & 49.60          & 27.67          & 25.76          & 39.69          & 37.92          & 21.22          & 18.63          & 12.28                         & 9.35          & 9.22           & 5.23          & 13.34          & 20.52                    \\
\quad+ knowledge retrieval \cite{cao2021knowledge}      & 51.32          & 27.14          & 25.69          & 41.23          & 38.86          & 23.25          & 21.15          & 13.59                         & \textbf{9.84}          & 9.24           & 5.51          & 13.89          & 21.30                    \\ \hline
% Ours(w/o VQA 2.0 knowledge)  & 58.97          & 42.98          & 36.11          & 48.27          & 45.93          & 34.70          & 29.33          & 16.25                         & 8.73 & \textbf{10.58} & 6.17          & \textbf{18.88} & 26.71                    \\
MuKEA & \textbf{59.12} & \textbf{44.88} & \textbf{37.36} & \textbf{52.47} & \textbf{48.08} & \textbf{35.63} & \textbf{31.61} & \textbf{17.62}                & 6.14          & \textbf{9.85}           & \textbf{6.22} & \textbf{18.28}          & \textbf{27.38}           \\ \hline
\end{tabular}}
\caption{State-of-the-art comparison on KRVQA dataset. The numbers in the third row mean different types of questions.}
\label{sota2}
\end{table*}

% \subsection{Datasets and Evaluation Metrics}
\noindent\textbf{Datasets and Evaluation Metrics.}\quad We conduct extensive experiments on two datasets: Outside Knowledge VQA (OK-VQA) \cite{marino2019ok} and Knowledge-Routed VQA (KRVQA) \cite{cao2021knowledge}. 
% are crowd-sourced from Amazon Mechanical Turkers, leading to 
OK-VQA contains more than 14,000 questions that cover a variety of 10 knowledge categories. It's diverse and challenging since all questions are human-annotated without fixed question templates or knowledge bases, which require exploring a wide range of open-ended knowledge resource. We evaluate the performance by the standard VQA evaluation metric \cite{antol2015vqa}. KRVQA \cite{cao2021knowledge} is to date the largest knowledge-based VQA dataset. It evaluates the multi-step reasoning ability of the models based on external knowledge. We use top-1 accuracy as in \cite{cao2021knowledge} for fair comparison.

% \subsection{Implementation Details}
\noindent\textbf{Implementation Details.}\quad %The inner feed-forward  network layer was set to 1024. The number of dimensions of entity and relation is set to 300. 
For all experiments, we train our model with PyTorch \cite{paszke2019pytorch}.
The softmax temperature $\tau$  in Eq. \ref{2} is set to 1.0. We used all the annotated answers in the train set to construct knowledge triplets. For the triplet ranking loss, we treat all samples in a batch with different answers from the positive samples as negative samples, the margin was set to 1.0. Our model is trained by AdamW \cite{loshchilov2018fixing} optimizer with 200 epochs, where the batch size is 256 and the learning rate is set to 1 × 10$^{-5}$ and  1 × 10$^{-4}$ in the pre-training and fine-tuning stage, respectively.

\subsection{Comparison with State-of-the-Art Methods}

\noindent\textbf{Comparison on OK-VQA:}\quad Table \ref{sota} shows the comparison results with state-of-the-art models, including knowledge graph based approaches \cite{DBLP:conf/ijcai/ZhuYWS0W20, DBLP:journals/pr/YuZWZHT20, garderes2020conceptbert, zheng2021knowledge}, unstructured-knowledge based approaches \cite{marino2019ok}, multi-source knowledge based hybrid approaches \cite{li2020boosting, marino2021krisp}, implicit knowledge based pre-training approaches \cite{tan2019lxmert, lu2019vilbert}, and the multimodal knowledge based approach \cite{zheng2021km4}. Meanwhile, we also compare with traditional VQA methods \cite{kim2018bilinear, ben2017mutan}.

Our model MuKEA consistently outperforms all the existing approaches and is superior to the state-of-the-art model \cite{zheng2021knowledge} remarkably by 3.35$\%$. Compared with most models following the `knowledge retrieve and read' pipeline and referring to fixed knowledge bases, our end-to-end model effectively avoids cascading error while benefits from human-focused diverse multimodal  knowledge. Moreover, our model greatly outperforms the pre-trained models by 10$\%$ since our model captures the question-centric and information-abstracted multimodal knowledge instead of simple vision and language co-occurrence `knowledge' in the pre-training framework. Though KM$^4$ leverages multimodal knowledge by correlating images with entities in the existing knowledge graph, it still lacks of knowledge with high-order complex relationships and is inferior to MuKEA by 11.27\%. %Compared with classification methods, our model predicts on a more diverse set of accumulated knowledge.  

% Compared with methods based on multimodal knowledge, our approach outperforms KM$^4$ \cite{zheng2021km4} by 11.27$\%$, the model KM$^4$ adds visual nodes to the existing knowledge graph without considering the multimodal high-order relation and fusion the implicit knowledge embedding for answer classification. By modeling complex visual and semantic causality and perform reasoning on accumulated knowledge, our model gains remarkable improvement.

% Compared with models pre-trained on VQA 2.0 dataset, ours outperforms KRISP(with mm pre.) by 3.69$\%$, indicating multimodal knowledge extraction task is more effective than VQA task.

\noindent\textbf{Comparison on KRVQA:}\quad In Table \ref{sota2}, we compare MuKEA with  traditional VQA models \cite{perez2018film, yu2018beyond, anderson2018bottom, yu2019deep} and the knowledge-based model \cite{cao2021knowledge}. `KB-not-related' questions only require basic visual knowledge while `KB-related' questions need 
factual knowledge from knowledge bases.

Our model consistently outperforms existing models and achieves a remarkable boost of 6.08\% on the overall metric over the best model \cite{anderson2018bottom}. It's worth to note that MuKEA obtains 7.81\% improvement on average over \cite{anderson2018bottom} on the `KB-not-related' questions, which indicates that even the vision-only questions require multimodal commonsense to bridge the low-level visual content and high-level semantics. MuKEA is inferior to some models on  two-step reasoning type 3 questions since that the answers of these questions are mostly relations while the accumulated and predicted tail entities of MuKEA are fact entities in most cases. %\cite{cao2021knowledge} retrieves on a finite knowledge set provided by the dataset, our model accumulates knowledge in an open-ended way and avoids cascading error.

\begin{table}[t]
\setlength{\tabcolsep}{3mm}{
\begin{tabular}{l|l}
\hline
\textbf{Method}                                 & \textbf{Accuracy}      \\ \hline
1.\quad MuKEA (full model)                              & \textbf{42.59}                  \\ \hline
\textbf{Ablation of Loss Function} \\
2.\quad w/o $\mathcal{L}_{\rm Tri}$                            & 41.35                  \\
3.\quad w/o $\mathcal{L}_{\rm Sem}$                    & 42.06                  \\
4.\quad w/o $\mathcal{L}_{\rm Tri}$ \&  $\mathcal{L}_{\rm Sem}$       & 40.84                      \\
5.\quad w/o $\mathcal{L}_{\rm TransE}$                               & 24.50                  \\ \hline
\textbf{Ablation of Triplet Representation}  \\
6.\quad head entity w/ soft-attention                       & 40.67                \\
7.\quad relation w/ self-attention                      & 40.79                  \\
8.\quad tail entity w/ GloVe                               & 41.42                  \\ \hline
\textbf{Ablation of Triplet Structure}              \\
9.\quad w/o $h$                     & 39.83                      \\ 
10.\: w/o $r$                    & 39.40                     \\ \hline
\textbf{Ablation of Knowledge Source}              \\
11.\: w/o VQA 2.0 knowledge                      & 36.35                      \\ 
12.\: w/o OK-VQA knowledge                    & 27.20                      \\ \hline
\textbf{Ablation of Pre-training Knowledge}             \\
13.\: w/o LXMERT pre-training                    & 33.52                      \\ \hline
\end{tabular}}
\caption{Ablation of key components in MuKEA on OK-VQA.}
\label{ablation}
\end{table}

\subsection{Ablation Study}
In Table \ref{ablation}, we evaluate the contribution of knowledge learning  losses, knowledge extraction schema, and knowledge accumulation strategy in MuKEA on the OK-VQA dataset. (1) In models `2-5', we evaluate the \textbf{effect of each loss function} on the performance. The accuracy of removing $\mathcal{L}_{\rm Tri}$ and $\mathcal{L}_{\rm Sem}$ respectively decreases by 1.24$\%$ and 0.53$\%$ while removing $\mathcal{L}_{\rm TransE}$ results in a significant decrease in model `5'. Because $\mathcal{L}_{\rm TransE}$ preserves the embedding structure of the whole triplets in our multimodal knowledge base, which has greater impact than $\mathcal{L}_{\rm Tri}$ and $\mathcal{L}_{\rm Sem}$. Model `4' results in a further decrease compared with `2' and `3', which indicates the complementary benefits of $\mathcal{L}_{\rm Tri}$ and $\mathcal{L}_{\rm Sem}$. (2) In models `6-8', we assess the \textbf{influence of triplet extraction methods}. For head entity extraction, we replace Gumbel-Softmax with soft attention in `6' and the performance drops by 1.92$\%$. It's because that the head entity  derived from LXMERT already contains object-centric contextual semantics for complex questions while directly fusing object features together introduces unexpected noise. Similarly, we apply self-attention over all the output tokens of LXMEART to represent the relation in `7' and the accuracy decreases by 1.80$\%$ compared with using [CLS] token, which benefits from the the pre-training classification task to contain highly-correlated multimodal information. Furthermore, we utilize GloVe \cite{pennington2014glove} to represent the tail entity in `8', resulting in a 1.17$\%$ accuracy drop because of the heterogeneous gap between fixed word embeddings and multimodal triplet representations. (3) In models `9-10', we prove the \textbf{importance of triplet structure}.  We remove the head entity and the tail entity respectively. The performance drops 2.76\% and 3.19\% accordingly, which proves the effectiveness of our triplet-based knowledge organization structure. (4) In models `11-12', we prove the \textbf{importance of pre-training and fine-tuning strategy for knowledge accumulation}. It's obvious that without either of the two process, the performance decreases remarkably. Though the basic knowledge in VQA 2.0 is less influential than the domain-specific knowledge in OK-VQA, the two works together in a curriculum learning mode achieves the best performance. (5) In model `13', we further test the \textbf{influence of prior knowledge accumulated in the pre-trained LXMERT}. The accuracy drops 9.07$\%$ without pre-training since both the head entity and the relation representations rely on the contextual information from the pre-trained knowledge.

\begin{table}[t]
\centering
\setlength{\tabcolsep}{1mm}{
\begin{tabular}{l|c|c|c}
\hline
\multirow{2}{*}{\textbf{Method}} & \multicolumn{3}{c}{\textbf{Failure subset}}     \\ \cline{2-4} 
                                 & \textbf{MUTAN + AN$^*$} & \textbf{Mucko$^*$} & \textbf{KRISP$^*$} \\ \hline
MuKEA                             & 40.09               & 40.06          & 40.46          \\ \hline
\end{tabular}
\\[3pt](a)\\[3pt]
\begin{tabular}{l|c}
\hline
\multirow{2}{*}{\textbf{Method}} & \textbf{Failure subset} \\ \cline{2-2} 
                                 & \textbf{MuKEA}                \\ \hline
MUTAN + AN$^*$                       & 26.45                        \\ 
Mucko$^*$                            & 27.68                        \\ 
KRISP$^*$                            & 27.68                        \\ \hline
\end{tabular}}
\\[3pt](b)\\
\caption{MuKEA accuracy on the failure subset of KB-based models (a) and vice versa (b). * indicates the model is re-implemented.}
\label{mispredict}
\end{table}

\subsection{Knowledge Complementary Analysis}
To prove the complementary benefits of our multimodal knowledge with existing knowledge bases, we conduct two experiments on the OK-VQA dataset: (1) performance of MuKEA and existing models on mutual failure cases, and (2) performance of ensemble models of MuKEA and existing models. Here we test on three typical KB-based models: MUTAN + AN \cite{marino2019ok} on unstructured Wikipedia, Mucko \cite{DBLP:conf/ijcai/ZhuYWS0W20} on structured ConceptNet, and KRISP \cite{marino2021krisp} on multiple knowledge bases. We re-implemented these models for fair comparison on the same subset. 

Table \ref{mispredict} shows the performance of MuKEA on the failure OK-VQA test subset of the above three models and vice versa. MuKEA consistently achieves over 40\% accuracy on all the failure cases of the KB-based models (Table \ref{mispredict}(a)). Meanwhile, KB-based models obtain over 26\% accuracy on questions difficult for MuKEA (Table \ref{mispredict}(b)). It proves that multimodal knowledge and existing KB knowledge respectively covers different types of open-ended questions.   

We further assemble MuKEA with three models respectively: if the difference of the top 2 minimum distances predicted by Eq. \ref{8} is larger than a threshold $m$ ($m=0.07$), we select the predicted answer of MuKEA, otherwise, selecting the other. In Table \ref{jicheng}, the baseline models are respectively improved by 9.96$\%$, 8.80$\%$ and 5.73$\%$ after model ensemble. We also present the oracle setting that takes the accurate prediction from either of the models as the answer. The oracle performance obtains  significant improvement, which further proves the complementary benefits of multimodal knowledge and existing knowledge bases.

\begin{figure*}[t]
\setlength{\abovecaptionskip}{-2pt}
\setlength{\belowcaptionskip}{-5pt}
\begin{center}
\includegraphics[width=1.0\linewidth]{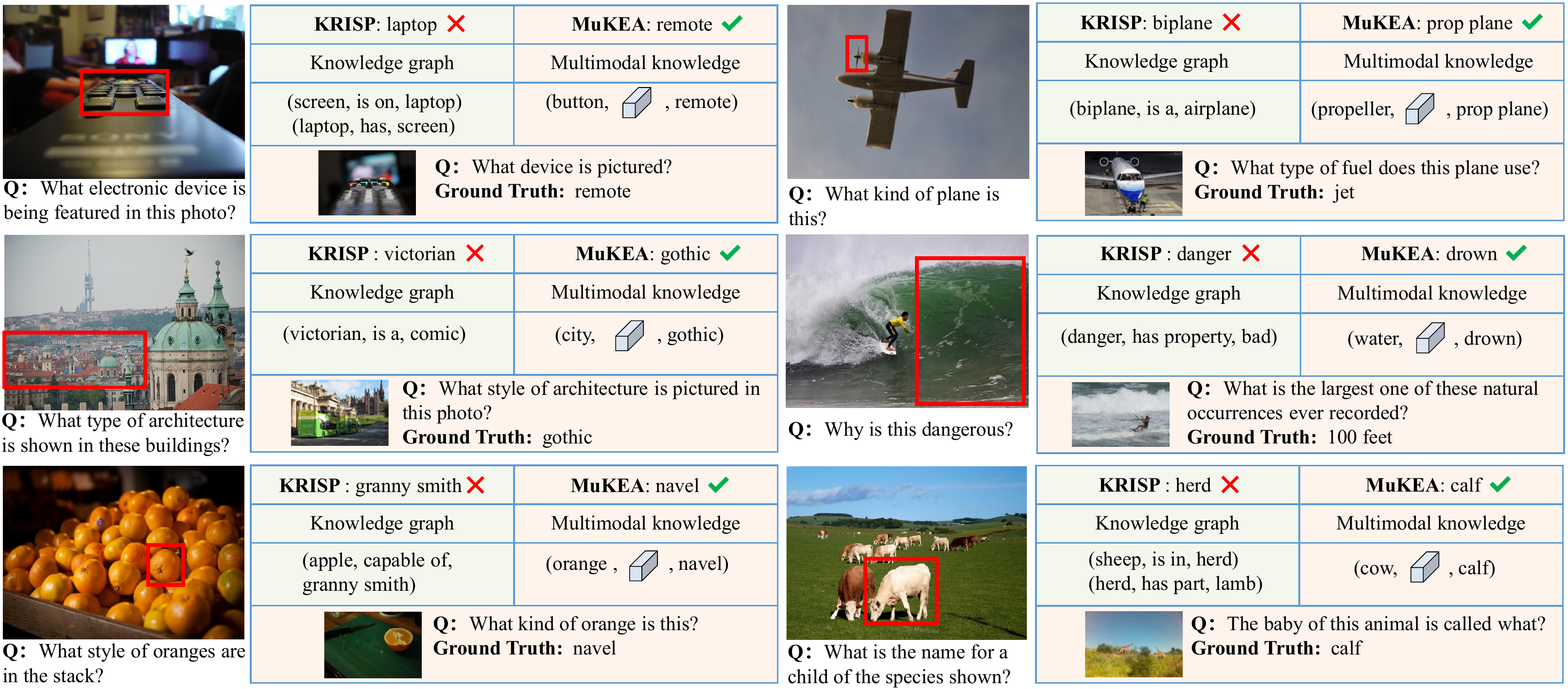}
\end{center}
\caption{Visualization of the predicted answers and supporting knowledge of KRISP (green) and MuKEA (pink). For MuKEA, the red box in the image shows the head entity ($\alpha_i$ in Eq. \ref{2}). The bottom VQA training sample, which has the nearest relation embedding with the test sample, shows the scenario that accumulates relational knowledge supporting the current inference. The answer is shown in the tail.}
\label{fig:qua}
\end{figure*}

\begin{table}[t]
\setlength{\belowcaptionskip}{-10pt}
\centering
\setlength{\tabcolsep}{1mm}{
\begin{tabular}{l|c}
\hline
\textbf{Method}                & \textbf{Accuracy} \\ \hline
MuKEA                             & 42.59              \\ \hline
MUTAN + AN$^*$                    & 25.43             \\
MuKEA + (MUTAN + AN$^*$)        & 35.39             \\
MuKEA + (MUTAN + AN$^*$) oracle & 43.64             \\ \hline
Mucko$^*$                      & 27.17             \\
MuKEA + Mucko$^*$               & 35.97             \\
MuKEA + Mucko$^*$ oracle        & 44.84             \\ \hline
KRISP$^*$                      & 32.02             \\
MuKEA + KRISP$^*$               & 37.75             \\
MuKEA + KRISP$^*$ oracle        & 47.15             \\ \hline
\end{tabular}}
\caption{Performance of model ensemble on OK-VQA.}
\label{jicheng}
\end{table}

\begin{table}[t]
\setlength{\belowcaptionskip}{-10pt}
\centering
\setlength{\tabcolsep}{1mm}{
\begin{tabular}{l|c|c}
\hline
\textbf{Method}                     & \textbf{Accuracy} &  \textbf{mAccuracy} \\ \hline
KRISP$^*$                      & 32.31    & 26.91     \\
MuKEA & 42.59    & 35.42     \\ \hline
\end{tabular}}
\caption{Long-tail analysis on OK-VQA dataset.}
\label{longtail}
\end{table}

\subsection{Long-Tail Analysis}
To prove the model's generalization ability on the rare answers while not overfitting on the `head' ones, we propose a new unbiased metric mean Accuracy (\textbf{mAccuracy}) to fairly evaluate the performance on the long-tail distributed answers. Inspired by the unbiased metric in scene graph generation \cite{chen2019knowledge, tang2019learning}, mAccuracy calculates the accuracy for each unique answer separately and average for all the answers. We compare MuKEA with KRISP, which demonstrates its great generalization ability by referring to multiple knowledge sources. In Table  \ref{longtail}, our model greatly outperforms KRISP by 8.51\% on mAccuracy, which proves the strong generalization ability of the multimodal knowledge on the long-tail knowledge without sacrificing the accuracy of frequent referred knowledge. 

%Next, we try to see whether our model performs well on the `long tail' of OK-VQA. OK-VQA itself is built as a long-tail dataset, specifically rejecting answers that appear too many times to avoid models overfitting to the answer vocabulary, we use a mAccuracy metric to study whether our method does better on rare answers. mAccuracy calculates Accuracy for each class independently, and then average the results. We set the answer with the most occurrences among the 10 annotated answers as the answer category.

%In Table \ref{longtail} we show metrics on our model versus the KRISP \cite{marino2021krisp}, KRISP integrates implicit and symbolic knowledge and generalizes well to the long-tail. On mAccuracy metric our model outperforms 8.51$\%$ by KRISP, by accumulating both out-domain and in-domain knowledge to form a neural multimodal knowledge base, our model increases the diversity of answers. These results indicate the ability of multimodal knowledge to generalize the long-tail. 

\subsection{Qualitative Analysis}
From case study in Figure \ref{fig:qua}, we conclude that our model is interpretable by visualizing the predicted multimodal knowledge triplets: (1) \textbf{MuKEA captures instantiated knowledge beneficial for object understanding.} The examples in the first row indicate that MuKEA captures the complex knowledge between the object appearance and the object-centric facts. The accumulated knowledge is in the form of an entire triplet (left) or just the inexpressible relation (right). (2) \textbf{MuKEA contains multi-object involved complex knowledge beneficial for scene understanding.} In the second row, MuKEA is capable to correlate the visual content of groups of buildings with the abstract city style `gothic'. (3) \textbf{MuKEA avoids the cascading error by directly reasoning on knowledge embeddings.} Existing models generally first detect object labels to retrieve relevant knowledge, which introduces unexpected noise with false labels. MuKEA has the advantage of adopting the semantic-rich embeddings to represent the knowledge and reason about the answer in an end-to-end mode. 

% understand how the multimodal knowledge might be helping our model to answer questions
%(1) \textbf{MuKEA leverage multimodal knowledge to help with fine-grained visual reasoning}. In the first three examples, our model correctly answers the name of the device, type of plane and style of architecture by multimodal knowledge. (2) \textbf{MuKEA extract visual and semantic causality to represent high-order relations}. In the middle right example, our model successfully established a connection between water and drown. (3) \textbf{MuKEA perform multimodal reasoning to get rid of the cascading error in existing `knowledge retrieve and read' pipelin}. In the last two examples, since KRISP uses the wrong semantic labels ($e.g.$ apple and sheep) for knowledge retrieval, resulting in cascading errors.
% our model focuses on key areas in images with dense objects, and utilizes visual embedding for knowledge reasoning. 

\subsection{Limitation Analysis}
\begin{figure}[t]
\begin{center}
\includegraphics[width=0.7\linewidth]{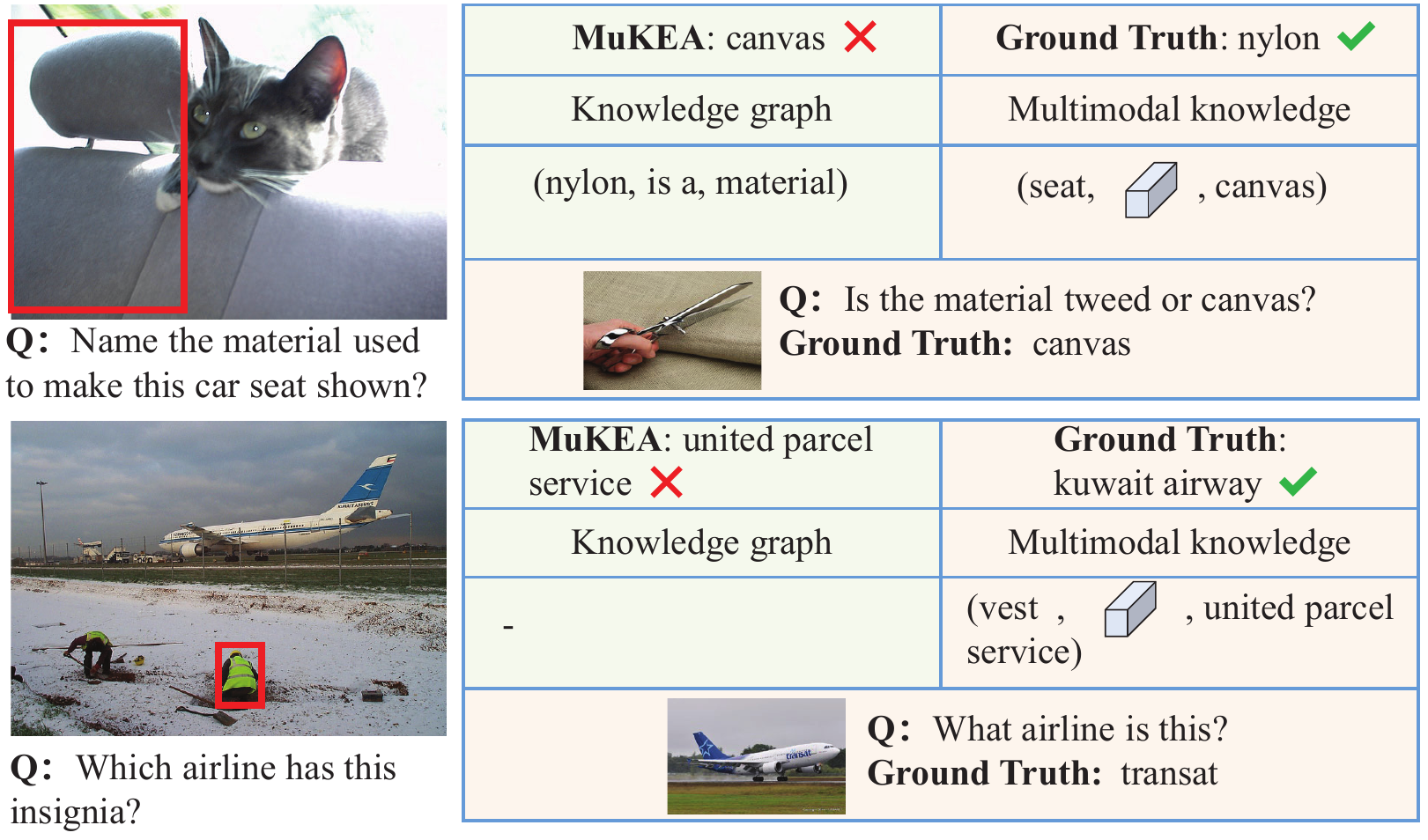}
\end{center}
\vspace{-15pt}
\caption{Representative failure cases of MuKEA on OK-VQA.} 
\vspace{-16pt}
% Showing predictions by our model and the GT(ground truth). The meanings of symbols are the same as Figure \ref{fig:qua}}
\label{fig:err}
\end{figure}

MuKEA fails mostly in the following cases (Figure \ref{fig:err}): (1) %\textbf{Limited multimodal knowledge}. 
MuKEA lacks adequate multimodal knowledge, such as the knowledge to distinguish \textit{nylon} and \textit{canvas},
% in the first case
due to the limited VQA scenarios in the training stage.  (2) MuKEA fails in extracting some triplets. 
% Since the head entities and the relations are extracted in a totally unsupervised mode
Since the head entities and the relations are extracted in an unsupervised mode, vision-similar content causes attention deviation, such as the vest is incorrectly attended as the \textit{insignia}. The above problems need further research in accumulating  more comprehensive knowledge and evaluating the triplet extraction quality. We also test the MuKEA on VQA 2.0 with inferior results to some works since that questions in VQA 2.0 mainly rely on visual appearance clues instead of external knowledge. 
% Though the number of tail entities in the multimodal knowledge increases with the growth of training data, the inference time is not affected much when varying the knowledge size.  
% It indicates that the basic multimodal knowledge only is not enough to tackle vision-based VQA problems. Effectively combining traditional VQA models with multimodal knowledge may brings extra improvement.

%in-depth study can be : (1) Construction of large-scale multimodal commonsense knowledge graph. (2) Unsupervised multimodal knowledge triplet extraction. (3) Combining multimodal knowledge and existing knowledge graph to perform explicit knowledge reasoning. 

\section{Conclusion}
In this paper, we propose a novel framework for knowledge-based visual question answering, which focuses on multimodal knowledge extraction and accumulation instead of using external knowledge bases. We propose a novel schema to represent multimodal knowledge by an explicit triplet and three loss functions to learn the representations from coarse to fine. We adopt a pre-training and fine-tuning strategy to accumulate multimodal knowledge progressively. Our model outperforms state-of-the-art on KB-VQA datasets and advances recent research from the multimodal knowledge view. We prove the complementary to existing knowledge graph. How to effectively combine MuKEA with knowledge bases will be the future work.

\FloatBarrier
%%%%%%%%% REFERENCES
{\small
\bibliographystyle{ieee_fullname}
\bibliography{egbib}
}

\clearpage
\appendix
\renewcommand{\appendixname}{Appendix~\Alph{section}}
\setcounter{table}{0}
\setcounter{figure}{0}

\twocolumn[
\begin{@twocolumnfalse}
\section*{\centering{\large MuKEA: Multimodal Knowledge Extraction and Accumulation for Knowledge-based Visual
Question Answering \\[10pt] Supplementary Material \\[20pt]}}
\end{@twocolumnfalse}
]

\begin{abstract}
We provide additional materials to supplement our main submissions. In Section \ref{know_construct}, we introduce explicit multimodal knowledge construction, knowledge graph characteristics, application scenarios in detail, and provide extracted multimodal knowledge embeddings as off-the-shelf knowledge features to serve knowledge-based downstream tasks. Based on the knowledge graph constructed above, in Section \ref{know_acct} and \ref{know_zero}, we respectively introduce how MuKEA performs multimodal knowledge accumulation and complex reasoning. Then we compare the model size of MuKEA with pre-trained models and analyse the influence of multimodal knowledge base size on inference time respectively in Section \ref{param} and \ref{efficiency}, which proves that the inference time is not affected much when varying the knowledge size. In Section \ref{ensemble}, we study the effect of hyper-parameters in model ensemble corresponding to the knowledge complementary experiments. At last, in Section \ref{imple}, we introduce the implementation details about training.

% In Section \ref{imple}, we introduce the implementation details about training. In Section \ref{2.0}, we test the pre-trained MuKEA on VQA 2.0 \cite{goyal2017making}. In Section \ref{ensemble}, we study the effect of hyper-parameters in model ensemble. In Section \ref{param}, we compare the model size of MuKEA with pre-trained models. In Section \ref{efficiency}, we analyse the influence of multimodal knowledge base size on inference time. In Section \ref{visual}, we present more in-depth visualization analysis of our model.  
\end{abstract}

%%%%%%%%% BODY TEXT

\begin{figure*}[htb]
\begin{center}
\includegraphics[width=1.0\linewidth]{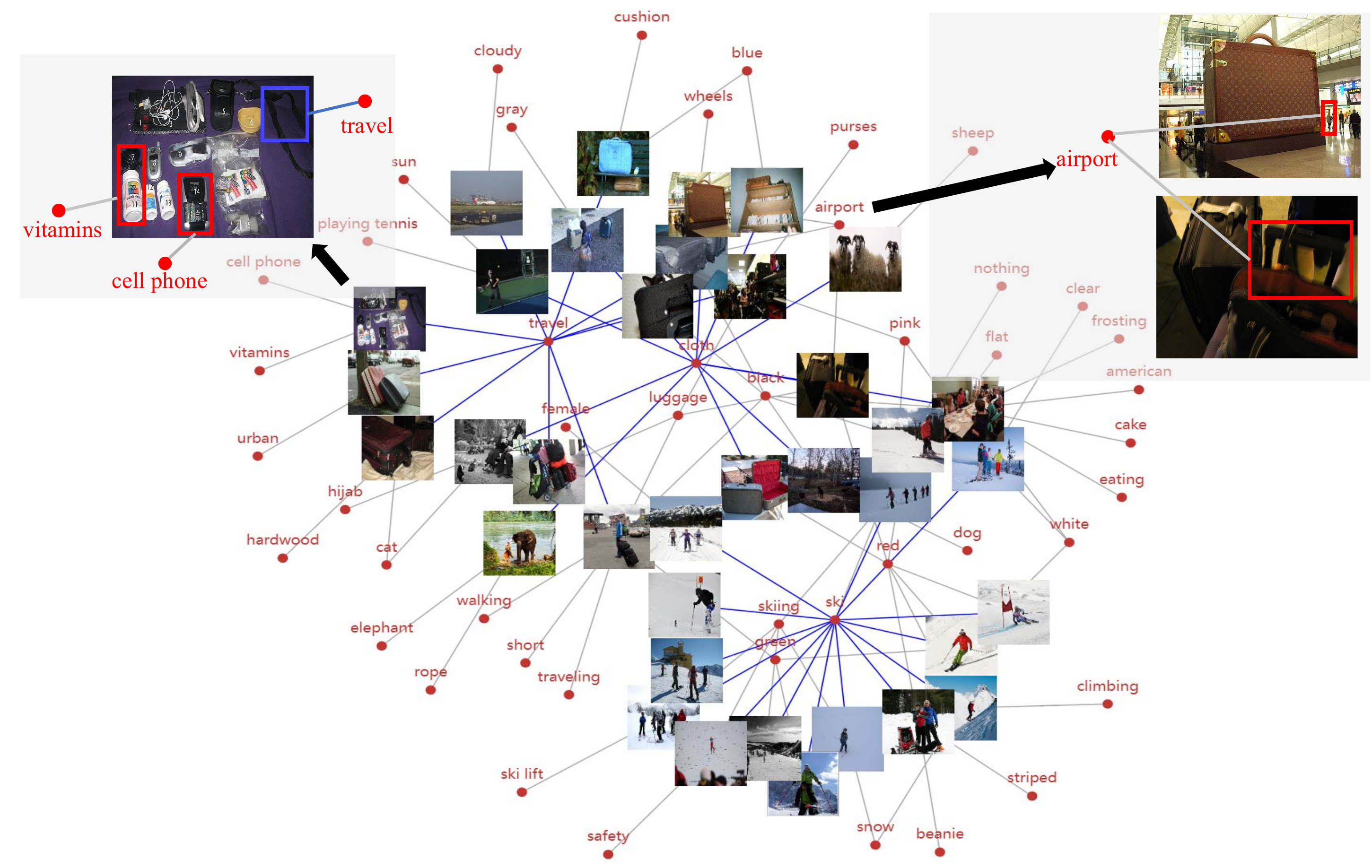}
\end{center}
\caption{Visualization of the accumulated knowledge graph. Gray lines and blue lines means knowledge accumulated in the VQA 2.0 and OK-VQA respectively. We show extra zoom-in examples for demonstration.}
\label{fig:acc2}
\end{figure*}

\begin{figure*}[t]
\setlength{\abovecaptionskip}{-2pt}
\setlength{\belowcaptionskip}{-2pt}
\begin{center}
\includegraphics[width=1.0\linewidth]{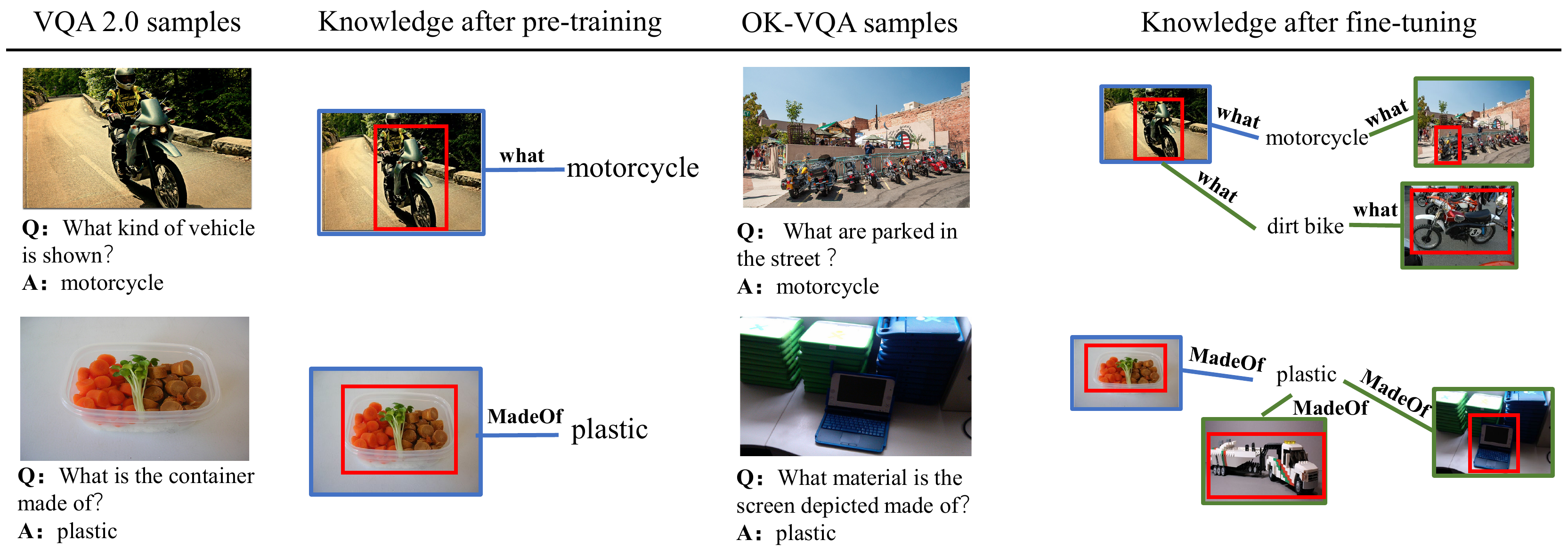}
\end{center}
\caption{Knowledge accumulation in the pre-training stage and fine-tuning stages.}
\label{fig:acc1}
\end{figure*}

\begin{figure*}[t]
% \vspace{-12pt}
% \setlength{\abovecaptionskip}{-5pt}
% \setlength{\belowcaptionskip}{-18pt}
\begin{center}
\includegraphics[width=1.0\linewidth]{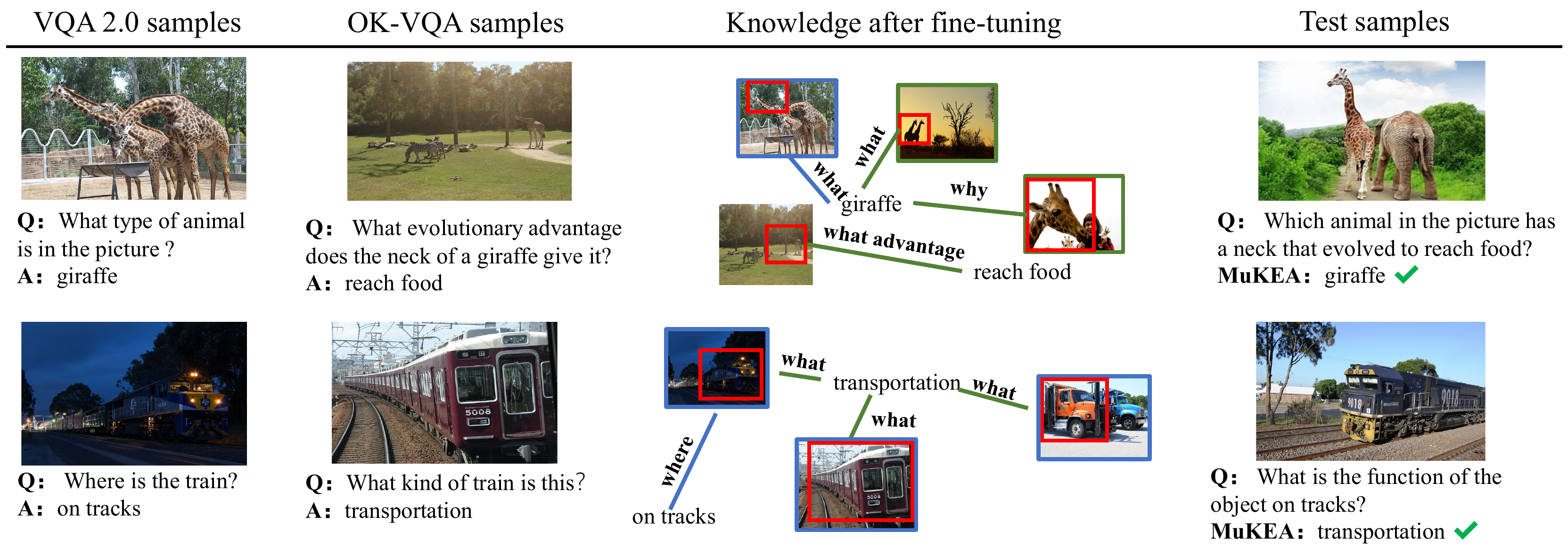}
\end{center}
\caption{Testing samples based on manually constructed questions in zero-shot setting.}
\label{fig:acc4}
\end{figure*}

\section{Multimodal Knowledge Construction and Related Applications}
\label{know_construct}
% After the pre-training and fine-tuning stage, we accumulate 218,135 multimodal knowledge triplets for knowledge retrieval and answer prediction. In Figure \ref{fig:acc2}, we select the samples with top-3 frequent answers in the OK-VQA train set and related samples in VQA 2.0, then visualize the sub-graph of our accumulated knowledge triplets. The example in the green box indicates that MukEA extracts different instantiated knowledge for the same image based on different regions in the image, which can help with answering questions in different types.
In Figure \ref{fig:acc2}, we show a 1-hop sub-graph of our accumulated knowledge triplets centered in top-3 frequent answers in the OK-VQA train set. To construct multimodal knowledge graph, firstly we store the extracted knowledge triplet $(h, r, t)$ from training data, where $h$ is the visual region in the image focused by the question, $t$ is the ground-truth answer, and $r$ is the embedding of implicit relation between $h$ and $t$. We only display the image corresponding to the $h$ and $t$ to explicitly show the structure of the accumulated multimodal knowledge graph. Then we merge all the tail entities with the same answer and merge all the head entities with the same image, while preserving object regions in images as shown in the example. After the pre-training and fine-tuning stages, we accumulate 218,135 multimodal knowledge triplets for knowledge graph construction.

We summarize the characteristics of our proposed multimodal knowledge graph as follows:
\begin{itemize}
    \item MuKEA extracts different instantiated knowledge for the same image based on different objects in the image. As shown in the left zoom-in example in Figure \ref{fig:acc2}, the same image connects both `cell phone' and `travel' since different objects in the same image related to different knowledge. On the contrary, existing multimodal knowledge graph \cite{wang2020richpedia} is a complement of general knowledge graphs with entity-referred images.
    \item The same concept is associated with different visual knowledge in different scenes. As shown in the right zoom-in example in Figure \ref{fig:acc2}, `airport' can correspond to different scenarios, such as airport hall or suitcases in airport. 
    \item Compared to existing knowledge graphs with pre-defined types of relations, relation in our proposed multimodal knowledge graph is extensible and supports retrieval as well.
    \item By correlating relevant knowledge, the knowledge graph is capable of supporting complex reasoning. In Section \ref{know_zero} we provide a detailed analysis. 
\end{itemize}

Furthermore, We propose the following potential application scenarios for using our multimodal knowledge graph:
\begin{itemize}
    \item Model-based knowledge search. MuKEA is capable of retrieving relevant knowledge for multimodal input. 
    \item Knowledge-based vision-language tasks, such as image caption, referring expression comprehension, vision-language navigation etc. 
    \item Explainable deep learning, especially in the legal, medical fields.
\end{itemize}

\textbf{The checkpoint of MuKEA, extracted multimodal knowledge graph, and off-the-shelf knowledge embeddings are available at \url{https://github.com/AndersonStra/MuKEA}}

\section{Analysis of Progress Knowledge Accumulation}
\label{know_acct}
From case study in Figure \ref{fig:acc1}, we illustrate how the basic visual knowledge in VQA 2.0 helps to learn more complex knowledge in OK-VQA: (1) In the first row, benefiting from the question about the appearance of motorcycle, MuKEA is capable to correlate the visual content of motorcycles with the answer in a multi-object scenario. (2) In the second row, benefiting from the prior knowledge of visual content with plastic materials, MuKEA has the advantage of focusing on the key region and obtaining more generalized representation for objects made of plastic.

\section{Zero-shot Analysis of Accumulated Multimodal Knowledge}
\label{know_zero}
In Figure \ref{fig:acc4}, we show that our model is capable to combine different accumulated knowledge to answer complex questions in the zero-shot setting. (1) In the test sample of the first row, we correlate `giraffe' with `evolution' through the manually constructed question. (2) In the test sample of the second row, we construct the question to correlate `track' and `transportation'. MuKEA performs correct prediction on both questions, which indicates that the accumulated multimodal knowledge can be applied to complex reasoning tasks in similar way as existing knowledge graphs.

\section{Model Size Analysis}
\label{param}
In Table \ref{table:size}, we compare the model size of MuKEA with pre-trained models \cite{su2019vl, lu2019vilbert, tan2019lxmert, marino2021krisp}. For MuKEA, we set the knowledge base size to the accumulated knowledge from VQA 2.0 \cite{goyal2017making} and OK-VQA \cite{marino2019ok}. The model size increases as the multimodal knowledge base size increases. Compared to ViLBERT, our model size only increases by 8.36\% with the performance boost by 11.24\%. The model size of KRISP is larger than ours by 86.78\%, but its performance is inferior to ours by 3.69\%. It indicates that our improvement is not from more parameters, but from the model structure.

\begin{table}[t]
\centering
\begin{tabular}{l|c}
\hline
Model   & Parameter \\ \hline
VL-BERT \cite{su2019vl} & 138.4M    \\
ViLBERT \cite{lu2019vilbert} & 218.9M    \\
LXMERT \cite{tan2019lxmert}  & 183.5M    \\ 
KRISP \cite{marino2021krisp} & 443.04M    \\ \hline
MuKEA   & 237.2M    \\ \hline
\end{tabular}
\caption{Comparison of model size.}
\label{table:size}
\end{table}

\begin{table*}[t]
\centering
\begin{tabular}{l|ccc}
\hline
\textbf{Knowledge Scale} & \textbf{Inference Time(s)} & \textbf{Ranking Time(s)} & \textbf{Ranking/Inference} \\ \hline
1000                     & 65.1431                         & 0.0026                & 0.0040\%                \\ \hline
10000                    & 65.1459                         & 0.0054                & 0.0083\%                \\ \hline
100000                   & 67.3542                         & 0.0064                & 0.0095\%                \\ \hline
\end{tabular}
\caption{Inference time and ranking time comparison based on different scale of knowledge base on OK-VQA.}
\label{table:eff}
\end{table*}

\section{Efficiency Analysis}
\label{efficiency}
To verify that MuKEA strikes a good balance between efficiency and effectiveness , we compare the inference time and ranking time separately based on different scale of multimodal knowledge base. We test on OK-VQA dataset \cite{marino2019ok}, which contains 5,046 samples for testing. Knowledge scale means the number of accumulated multimodal knowledge triplets. Inference time means the time spent on predicting over the entire test set. Ranking time means the time it takes to calculate the similarity with all tail entities in the knowledge base and rank the candidate tail entity.

Table \ref{table:eff} shows that although the ranking time is positively correlated with the size of knowledge base. It is relatively faster in the total inference time (less than 0.01\%, as shown in the column of `Ranking/Inference'). Since the GPU uses threads to process matrix multiplication in parallel, ranking time is not linearly related to the size of knowledge base. Our model still has good efficiency based on large-scale multimodal knowledge base.

\begin{figure}[t]
\begin{center}
\includegraphics[width=0.9\linewidth]{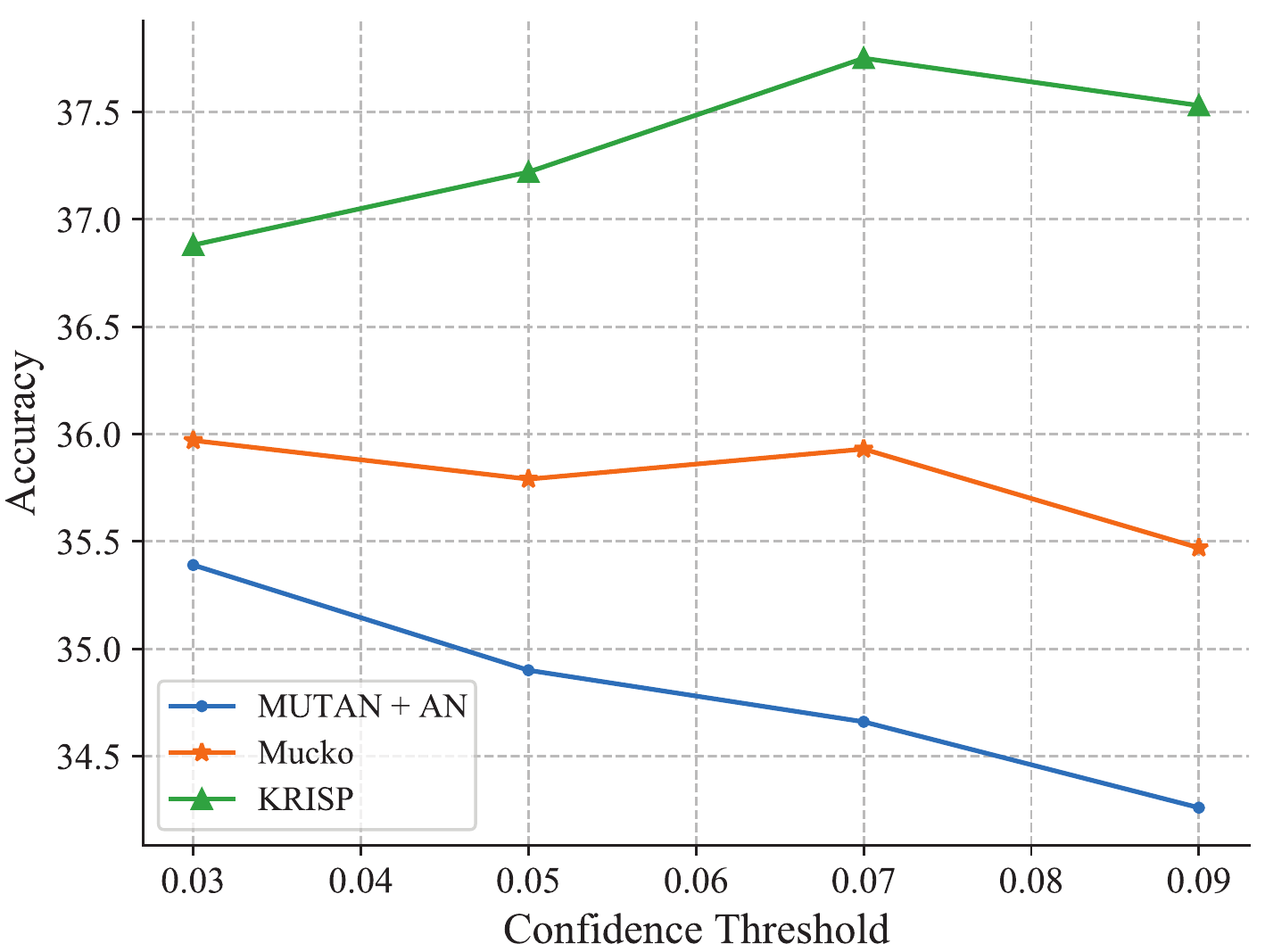}
\end{center}
\caption{Quantitative study of the confidence threshold $\tau_t$.}
\label{fig:para}
\end{figure}

\section{Effect of Ensemble Hyper-parameters}
\label{ensemble}
To verify the robustness of our ensemble model, we report the results of different threshold $m$ for model ensemble in Figure \ref{fig:para}. Although we propose a simple method based on confidence to perform model ensemble, we can find that the performance of ensemble model remains stable as the hyper-parameters change in a reasonable range. The threshold $m$ is set in the range of 0.03 and 0.09, and the performance of the ensemble models varies in the range of 34.24\% to 35.39\%, 35.49\% to 35.97\%, 36.88\% to 37.79\% respectively. How to effectively combine MuKEA with knowledge bases will be the future work.

\section{Analysis on VQA 2.0}
\label{2.0}
To prove the generalization ability of our method, we compare our model with state-of-the-art models on the VQA 2.0 dataset \cite{goyal2017making}, which  requires models to understand the visible content instead of incorporating external knowledge. All questions in VQA 2.0 are divided into three categories: \textit{Yes/No}, \textit{Number}, and \textit{Other}. Since our model is pre-trained on \textit{Other} type questions for accumulating basic multimodal knowledge, we only keep \textit{Other} type questions for comparison. 

\begin{figure}[t]
\begin{center}
\includegraphics[width=1.0\linewidth]{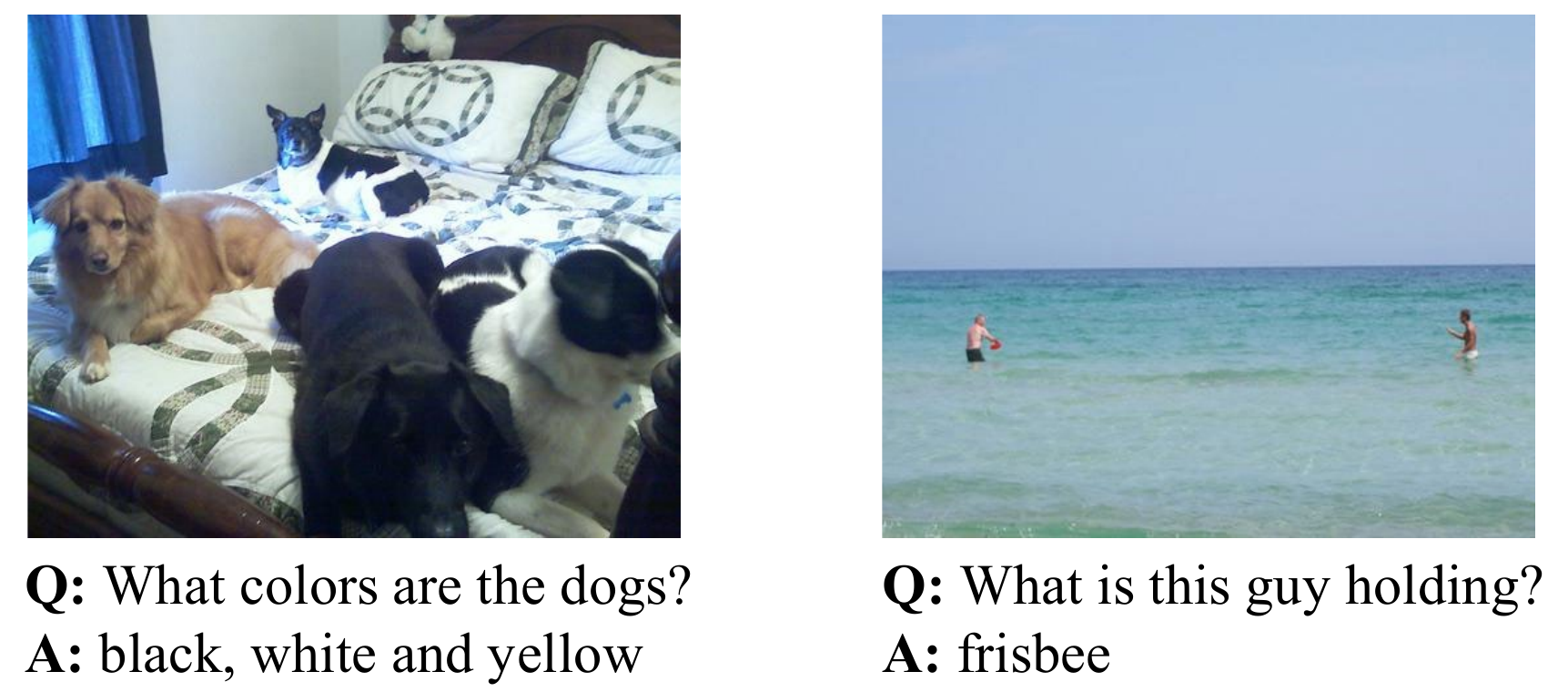}
\end{center}
\caption{samples in the VQA 2.0 dataset.}
\label{vqa_supp}
\end{figure}

Table \ref{VQA} shows that our model achieves comparable results compared with state-of-the-art models. This is mainly due to the following reasons: (1) VQA 2.0 mainly relies on visual appearance clues instead of external knowledge. As shown in Figure \ref{vqa_supp}, the example on the left requires the model to sense colors in multiple regions, while the example on the right requires the model to accurately detect object in the target region. (2) Existing models takes the head answers as the candidate answers, we accumulate multimodal knowledge on the whole dataset to ensure the diversity of answers, which is 10 times larger than the candidate answer set.

\begin{table}[t]
\centering
\begin{tabular}{lcc}
\hline
\multirow{2}{*}{Method}        & test-dev & test-std \\  \cmidrule(lr){2-2} \cmidrule(lr){3-3}
                              & Other    & Other    \\ \hline
\multicolumn{1}{l|}{MLB \cite{kim2016hadamard}}       & 56.34    & -        \\
\multicolumn{1}{l|}{MUTAN \cite{ben2017mutan}}     & 56.50    & -        \\
\multicolumn{1}{l|}{DCN \cite{nguyen2018improved}}       & 57.44    & 56.83    \\
\multicolumn{1}{l|}{DA-NTN \cite{bai2018deep}}    & 57.92    & -        \\
\multicolumn{1}{l|}{Counting \cite{zhang2018learning}}  & 58.97    &          \\
\multicolumn{1}{l|}{BLOCK \cite{ben2019block}}     & 58.51    & 58.79    \\
\multicolumn{1}{l|}{UpDn \cite{anderson2018bottom}}      & 56.05    & 56.26    \\
\multicolumn{1}{l|}{CGN \cite{norcliffe2018learning}}       & -        & 56.22    \\
\multicolumn{1}{l|}{CRA-Net \cite{peng2019cra}}   & 59.08    & 59.42    \\
\multicolumn{1}{l|}{MRA-Net \cite{peng2020mra}}   & 59.46    & 59.86    \\
\multicolumn{1}{l|}{SceneGCN \cite{9190771}}  & 57.77    & 57.89    \\
\multicolumn{1}{l|}{TRN+UpDn \cite{han2020interpretable}}  & 57.44    & -        \\
\multicolumn{1}{l|}{MuRel \cite{cadene2019murel}}     & 57.85    & -        \\
\multicolumn{1}{l|}{VCTREE+HL \cite{tang2019learning}} & 59.11    & 59.34    \\
\multicolumn{1}{l|}{LENA \cite{han2021focal}}      & 59.52    & 59.87    \\ \hline
\multicolumn{1}{l|}{Ours}      & 57.45    & 57.84    \\ \hline
\end{tabular}
\caption{Comparison on $Other$ split of VQA 2.0 dataset.}
\label{VQA}
\end{table}

\section{Implementation Details}
\label{imple}
For all experiments, we implement our model on top of LXMERT-based-uncased \cite{wolf-etal-2020-transformers} with 2 NVIDIA V100 GPUs. We follow \cite{DBLP:conf/ijcai/ZhuYWS0W20} to use Faster R-CNN model \cite{ren2015faster} pre-trained by the bottom-up model\cite{anderson2018bottom} on the Visual Genome dataset \cite{krishna2017visual}. The dimension of inner feed-forward network layer before head entity and relation is set to 1024. The dimensions of entity and relation in multimodal knowledge triplet are set to 300. The parameters in the look-up table are initialized by uniform distribution.

\end{document}

%% file: math_commands.tex
\usepackage{amsmath,amsfonts,bm}

\def\eqref#1{equation~\ref{#1}}
\def\1{\bm{1}}

\def\va{{\bm{a}}}

\def\vh{{\bm{h}}}

\def\vr{{\bm{r}}}

\def\vt{{\bm{t}}}

\def\mA{{\bm{A}}}

\def\mW{{\bm{W}}}

\DeclareMathAlphabet{\mathsfit}{\encodingdefault}{\sfdefault}{m}{sl}
\SetMathAlphabet{\mathsfit}{bold}{\encodingdefault}{\sfdefault}{bx}{n}

\newcommand{\R}{\mathbb{R}}